\documentclass{article}
\usepackage{graphicx}
\usepackage{float}
\usepackage{vmargin}
\usepackage{authblk}
\usepackage{amsmath}
\usepackage{amsfonts}
\usepackage{amssymb}
\usepackage{relsize}
\usepackage{bbm}
\usepackage{bm}
\usepackage{hyperref}
\usepackage{cleveref}
\usepackage{multirow}
\usepackage{lipsum}
\usepackage{cancel}

\title{Geological Inference from Textual Data using Word Embeddings}
\author[1,*]{Nanmanas Linphrachaya}
\author[1,a]{Irving Gómez-Méndez}
\author[2,b]{Adil Siripatana}
\affil[1]{CMKL University}
\affil[2]{The University of Edinburgh}
\affil[*]{Corresponding author: \href{nanmanaslinphrachaya@gmail.com}{nanmanaslinphrachaya@gmail.com}}
\affil[a]{\href{gomendez.irving@gmail.com}{gomendez.irving@gmail.com}}
\affil[b]{\href{asiripat@ed.ac.uk}{asiripat@ed.ac.uk}}

\date{}

\begin{document}
\maketitle

\begin{abstract}
   This research explores the use of Natural Language Processing (NLP) techniques to locate geological resources, with a specific focus on industrial minerals. By using word embeddings trained with the GloVe model, we extract semantic relationships between target keywords and a corpus of geological texts. The text is filtered to retain only words with geographical significance, such as city names, which are then ranked by their cosine similarity to the target keyword. Dimensional reduction techniques, including Principal Component Analysis (PCA), Autoencoder, Variational Autoencoder (VAE), and VAE with Long Short-Term Memory (VAE-LSTM), are applied to enhance feature extraction and improve the accuracy of semantic relations.

    For benchmarking, we calculate the proximity between the ten cities most semantically related to the target keyword and identified mine locations using the haversine equation. The results demonstrate that combining NLP with dimensional reduction techniques provides meaningful insights into the spatial distribution of natural resources. Although the result shows to be in the same region as the supposed location, the accuracy has room for improvement.
\end{abstract}

\section{Introduction}
    The exploration and extraction of natural resources, particularly minerals, are vital for economic development and technological advancement. As demand for specific resources, such as lithium increases, the need for efficient and cost-effective methods of resource location becomes even more pressing. Traditional exploration techniques, including geological surveys and physical fieldwork, though reliable, are both resource-intensive and time-consuming. Recent advances in data-driven methodologies, particularly Natural Language Processing (NLP), offer promising alternatives to supplement these conventional methods, enabling researchers to get valuable insights from vast amounts of geological text data \cite{lawley2022geoscience,lawley2023nlp}.

    NLP is a subfield of artificial intelligence that is now developing rapidly but its application in geological research remains much to be explored. The ability of NLP to extract semantic meaning from text can provide new avenues for understanding resource distribution by linking geological terms with geographical locations. Recent studies have shown that word embedding techniques, such as GloVe (Global Vectors for Word Representation), can capture the relationships between words by analyzing their co-occurrences in large corpora, allowing for the identification of patterns that may be indicative of resource locations \cite{pennington2014glove,mikolov2013efficient,sarker2019machine,brunsting2016geotexttagger}. 
    In this research, we apply these techniques to identify potential locations of lithium deposits based on their semantic similarity to geological terms extracted from relevant literature.

    To enhance the precision of this approach, we employ several dimensional reduction techniques, including Principal Component Analysis (PCA), Autoencoders, and Variational Autoencoders (VAE).
    These techniques allow us to reduce the complexity of the high-dimensional word embeddings, ensuring that the most relevant features for resource location prediction are retained \cite{goodfellow2016deep,kingma2013vae}.
    
    On the other hand, as  pointed out by several authors (see \cite{khatawakar2025evolution,yao2018improved}, and the references therein.) Long Short-Term Memory (LSTM) networks have been successfully applied in the field of NLP, alleviating some limitations of basic recurrent neural networks (RNN). Thus, we also explore the use of VAE in conjunction with LSTM networks (VAE-LSTM) in this study. 

\subsection{Contribution}
    The goal of this research is to assess the potential of NLP-based methods for predicting resource locations and to compare the performance of different dimensional reduction techniques in improving the accuracy of these predictions. By automating the initial stages of resource exploration, we aim to provide a tool that can streamline the process of identifying promising locations for further investigation. Given the critical importance of lithium for the future of renewable energy technologies, our research offers timely insights into how data-driven techniques can complement traditional exploration methods.
\section{Material and methods}
\label{sec:Method}
\subsection{Data Source}

    This research relies on several key data sources to perform NLP and resource location prediction.
    First and foremost is geological text data. We used the dataset published by British Columbia Geology, which provides comprehensive information on geological formations, mineral occurrences and regional geological surveys. This dataset were used to train the GloVe model for word embedding analysis, based on the previous work of \cite{lawley2023nlp}, and the programming codes available therein.

    Another data that we used in this research is the list of cities of Simplemaps \cite{simplemaps}. A widely-used database containing information on city names, population, and geographical coordinates (latitude and longitude), as it is shown in \Cref{tab:cities}. This dataset includes over 200,000 cities, providing a thorough coverage of global urban areas. The city names extracted from the geological texts were cross-referenced with this dataset to ensure geographical consistency. The coordinates of the cities were later used to calculate distances from known lithium deposits, enabling a spatial comparison to validate the NLP predictions.

\begin{center}
\begin{table}
    \centering
    \begin{tabular}{|c|c|c|c|c|c|c|c|}
        \hline
        \textbf{city} & \textbf{city-ascii} & \textbf{lat} & \textbf{lng} & \textbf{country} & \textbf{iso2} & \textbf{iso3} & \textbf{admin-name} \\
        
        \hline
         Tokyo       & Tokyo       & 35.6897  & 139.6922  & Japan    & JP    & JPN  & Tōkyō    \\
        \hline
         Jakarta     & Jakarta     & -6.175   & 106.8275  & Indonesia & ID    & IDN  & Jakarta \\
        \hline
         Delhi       & Delhi       & 28.6100  & 77.2300   & India    & IN    & IND  & Delhi \\
         \hline
         Guangzhou   & Guangzhou   & 23.1300  & 113.2600  & China    & CN    & CHN  & Guangdong \\
         \hline
         Mumbai      & Mumbai      & 19.0761  & 72.8775   & India    & IN    & IND  & Mahārāshtra \\
         \hline
    \end{tabular}
    \caption{Example list of cities from Simplemaps.}
    \label{tab:cities}
\end{table}
\end{center}

\begin{center}
    \begin{figure}
        \centering
        \includegraphics[width=0.5\textwidth]{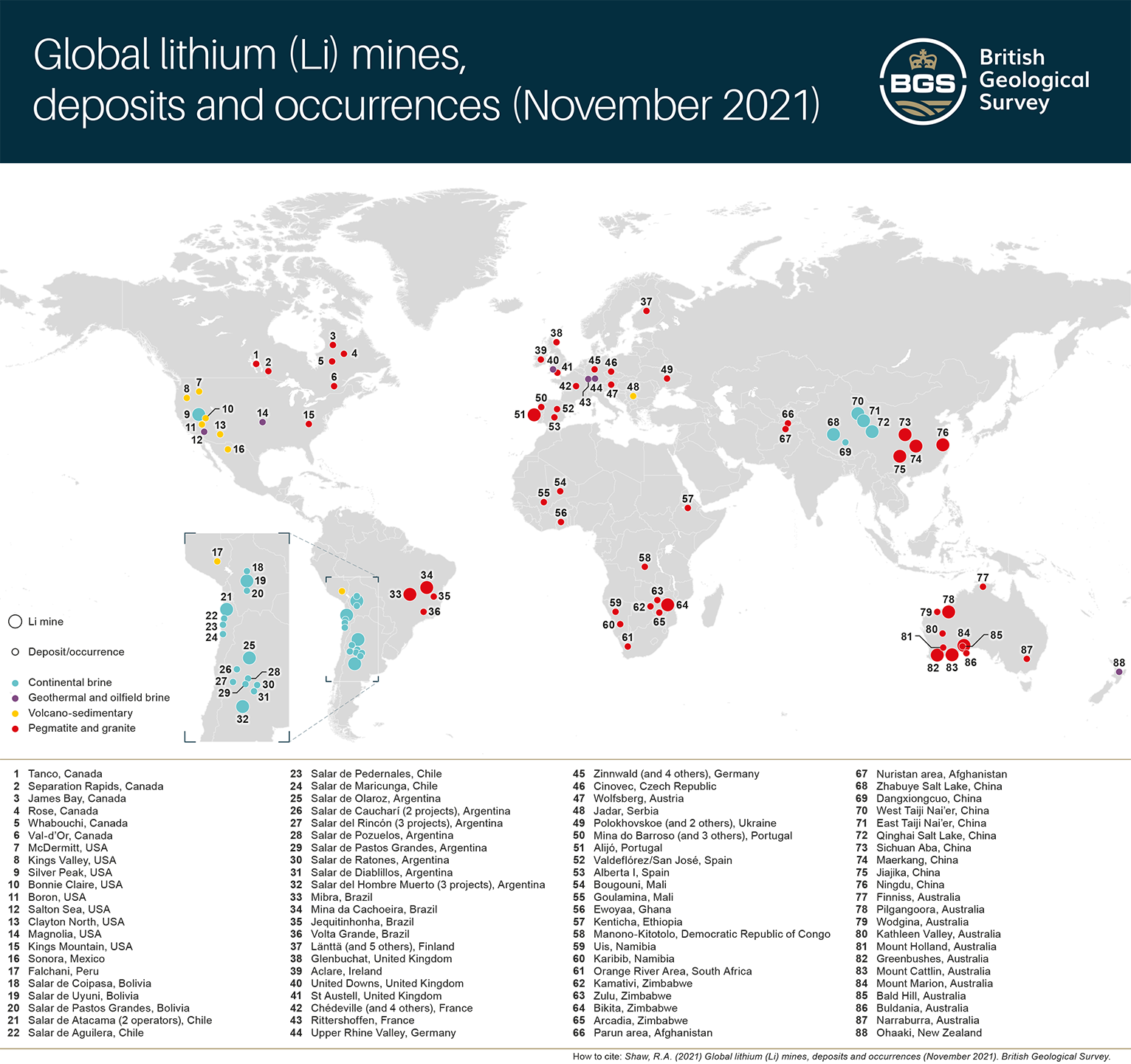}
        \caption{Map of lithium mines in the world, surveyed by British Geological Survey (BGS).}
        \label{fig:map_lithium_mines}
    \end{figure}
\end{center}

The final dataset used for benchmarking the result of the NLP analysis is Global Lithium Deposit Map provided by the British Geological Survey (BGS) \cite{bgs2023lithium} (see \Cref{fig:map_lithium_mines}). This map details known lithium deposits around the world, including both active mines and potential exploration sites. The locations of these lithium deposits were cross-referenced with the cities identified through semantic similarity analysis to evaluate the accuracy of the predictions. This map was chosen for its authoritative and up-to-date information on global lithium resources, providing a critical benchmark for validating the results of this research.

\begin{figure}
    \centering
    \includegraphics[width=1.2\textwidth]{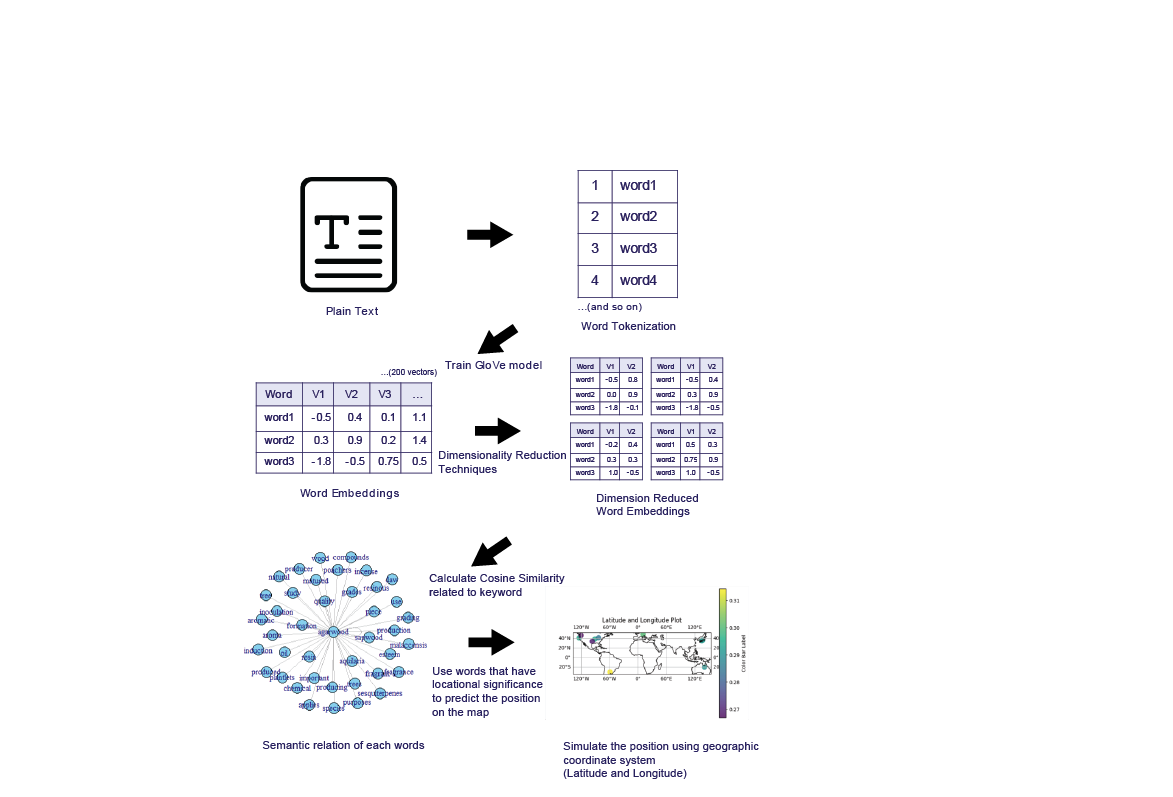}
    \caption{Overview of the methodology. Starting from pre-processing the text into tokens of text which will be used to train GloVe allowing them to be represented using word embeddings. These embeddings will be transformed based on dimensionality reduction techniques to filter the insignificant features. The correlation between the keyword and other words will be calculated using cosine similarity into scores which will indicate the locational significance word that will be used to predict the location of the selected keyword.}
    \label{fig:process_diagram}
\end{figure}

The rest of the process, outlined in \Cref{fig:process_diagram}, consists of the following steps: Text extraction, word embedding, dimensionality reduction, computation of cosine similarity, and calculation of haversine distance.

\subsection{Text Extraction}

The text processing for this research was based on the methodology outlined by \cite{lawley2023nlp} in their study on NLP applications in geoscience. Text processing was carried out using ‘‘tidyverse’’ \cite{wickham2019tidyverse}, ‘‘tidytext’’ \cite{silge2016tidytext}, and ‘‘sf’’ \cite{pebesma2018features} packages. The text processing followed the three core NLP tasks described by \cite{lawley2023nlp}: (1) tokenization, (2) removal of stop words, and (3) stemming.

This approach resulted in more meaningful word stems that aligned with the geoscience-specific GloVe model vocabulary used in the analysis \cite{lawley2022geoscience}.

\subsection{Word Embedding Process}

Word embedding transforms words from texts into dense vector representations that capture semantic relationships, creating a weighted matrix which allows similar words to be correlated with each other \cite{tarwani2017survey}. Unlike traditional one-hot encoding or TF-IDF approaches \cite{qaiser2018text}, embeddings like those produced by GloVe enable words to be placed in a continuous vector space, where semantic similarity is encoded through spatial proximity \cite{pennington2014glove}. This representation is particularly useful for identifying geological terms associated with mineral resources, as words occurring in similar geological contexts will have similar vector representations \cite{mikolov2013efficient}.

The GloVe model employed in this research was pre-trained on a large matrix of co-occurring words from extensive datasets, following the general approach outlined by \cite{lawley2023nlp}. Instead of using the original GloVe model, which was trained on six billion tokens from the Wikipedia 2014 and Gigawords datasets \cite{parker2011}, this study focused exclusively on texts from the BGS publications.

The GloVe model used here, as in \cite{pennington2014glove}, is based on the assumption that words appearing together frequently are semantically closer. While in \cite{lawley2023nlp} the authors re-trained their GloVe model on a broader collection of geoscientific documents from sources like Natural Resources Canada (NRCan) and various provincial geological surveys, this research re-trained the model only on BGS publications. This allows the creation of specialized embeddings for geological terms and concepts unique to British Columbia’s geological landscape.

Moreover, in contrast to the 300-dimensional vectors used by \cite{lawley2023nlp}, this study employed 200-dimensional vectors for each word, reducing the complexity while maintaining core semantic relationships. This methodology, adapted from \cite{mitchell2010,wieting2016,adi2017}, ensures that cities described by short or long geological texts are treated equivalently.

The final output of this adapted text processing pipeline is a data table containing 33,331 words used in training the model, along with a corresponding 200-dimensional vector that captures the geological characteristics associated with that word. These embeddings provide a robust numerical representation suitable for subsequent semantic analysis and cosine similarity calculations across various terms, enhancing the overall understanding of geological contexts.

After obtaining the table from training the GloVe model, a filtering process is applied to extract only meaningful words and city names. To achieve this, the list of words from the NLTK corpus \cite{bird2009nltk} and a list of cities obtained from Simplemaps \cite{simplemaps} were combined. This step ensures that only relevant geological terms and recognized city names are retained, excluding common, uninformative words.

After applying the filters, the final output consisted of 12,067 words, which included both city names and geological terms. This refined dataset improved the accuracy of subsequent semantic similarity calculations by removing irrelevant words and focusing on meaningful content.

\subsection{Dimensionality Reduction Techniques}

Dimensionality reduction is a critical step in this study to better understand the semantic relationship between words. Four techniques were employed: PCA, Autoencoder, VAE, and VAE-LSTM. The underlying structure for each method is presented in \Cref{tab:mse_loss}. Each technique contributes to improving feature extraction in different ways, which are outlined below.

\label{sec:Res}
\begin{center}
    \begin{table}
        \centering
        \begin{tabular}{
        |c|c|
        }
            \hline
            \textbf{Model} & \textbf{Structure} 
            \\
            \hline
             PCA & Linear projection to 2 dimensions 
             \\
             \hline
             Autoencoder & Hidden layers' dimension: 128,64,32,16,8; Latent dimension: 2 
             \\
             \hline
             VAE& Same as AE, with normality constraint for latent space 
             \\
             \hline
             VAE-LSTM & Same as VAE, with LSTM unit 
             \\
             \hline
        \end{tabular}
        \caption{Comparison of MSE loss on each dimensionality reduction technique}
        \label{tab:mse_loss}
    \end{table}
\end{center}

\paragraph{PCA:} PCA is a widely used statistical technique that transforms high-dimensional data into a lower-dimensional form by identifying the directions (principal components) along which the data varies the most. This method was used as an initial approach to reduce the word embeddings' dimensionality and eliminate redundant features.

PCA's main advantage is its simplicity. However, it lacks the capacity to capture complex, non-linear relationships within the data, which prompted the exploration of more sophisticated techniques like autoencoders.

\paragraph{Autoencoder:} An autoencoder is a type of neural network designed to learn a compressed representation (encoding) of the input data by forcing the network to reconstruct the input from the compressed form (decoding) \cite{hinton2006dimensionality}. By training the autoencoder to minimize the reconstruction error, the model learned to preserve key geological relationships in a reduced space. The compressed representations from the autoencoder allowed the model to efficiently handle large datasets while retaining meaningful patterns.

\begin{center}
    \begin{figure}
        \centering
        \includegraphics[width=\textwidth]{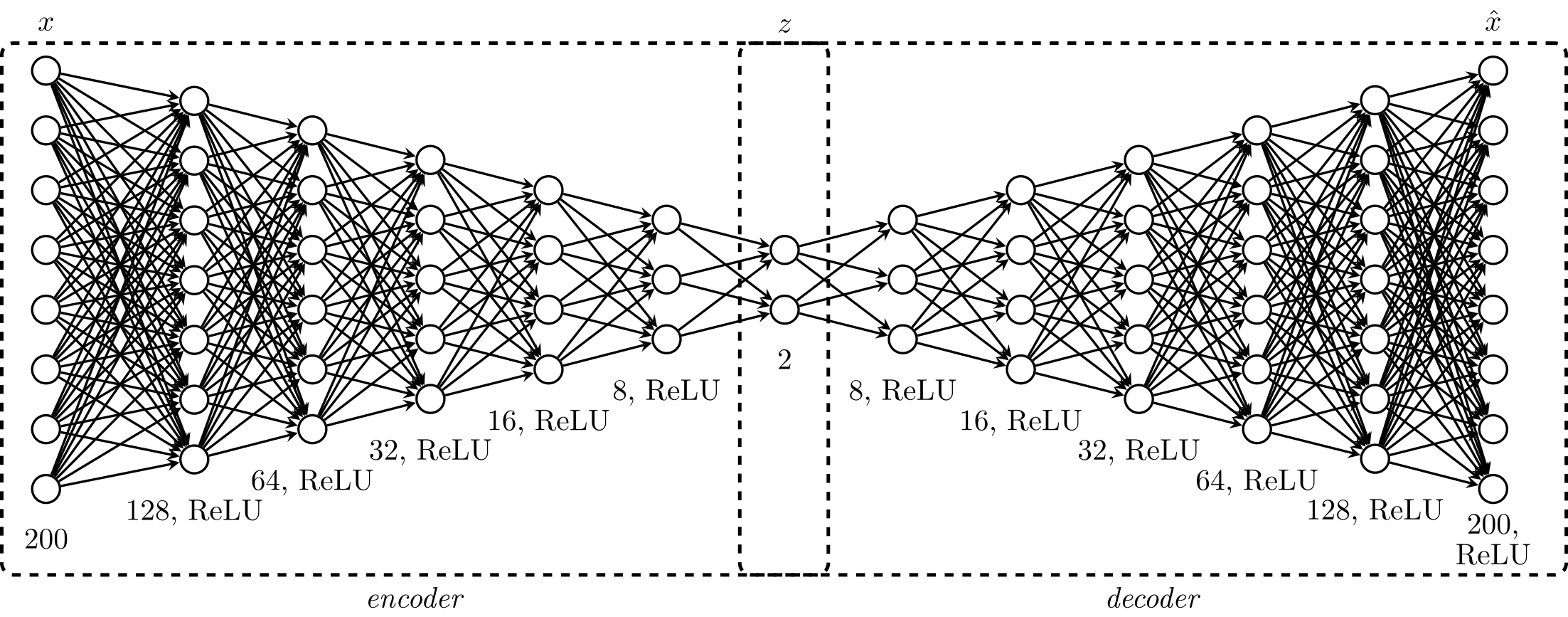}
        \caption{Graphical representation of the Autoencoder.}
        \label{fig:Autoencoder}
    \end{figure}
\end{center}

In this study, the autoencoder has five hidden layers for the encoding part, with dimensions 128, 64, 32, 16, and 8, followed by a 2-dimensional latent space, and using ReLU activation functions, while the decoding mirrors this structure, as represented in \Cref{fig:Autoencoder}.

\paragraph{VAE:} A VAE, introduced by \cite{kingma2013vae}, extends the idea of a regular autoencoder by introducing a probabilistic framework, encoding the input into a distribution rather than fixed points, and promoting smoothness and disentanglement within the latent space. This property allows for more robust feature extraction, particularly in capturing uncertainties and variations in geological descriptions. The advantage of VAEs lies in their ability to explore the latent space better, making them particularly useful for capturing subtle and complex relationships between words in geological texts. For an accessible introduction to variational inference and how VAEs achieve this, we recommend \cite{kingma2013vae}.

The VAE in this study used the same architecture as the autoencoder, adding the restriction that the variable in the latent space follows a normal distribution parameterized by its mean and variance, with these parameters optimized by minimizing both the reconstruction error and the Kullback-Leibler divergence.

\paragraph{VAE-LSTM:} To further enhance the model's capacity to capture sequential dependencies within the text, a VAE with LSTM architecture was employed. LSTMs are a form of RNN particularly suited for processing sequential data \cite{hochreiter1997lstm}. Integrating LSTMs into the VAE architecture helps to capture the semantical relationships between words and geological contexts, which is important for understanding sequences of descriptions commonly found in geological reports. Thus, VAE together with LSTM allows the model to capture not just the relationships between words but also the flow and context of geological descriptions, providing a richer semantic understanding.

\subsection{Cosine Similarity Calculation}

In this study, cosine similarity was used to identify cities most semantically related to a specific geological keyword. Cosine similarity provides a quantitative measure of similarity by calculating the cosine of the angle between two vectors, independent of their magnitudes, being commonly used in NLP tasks for assessing the similarity between word embeddings while using less computational power than other alternatives such as Euclidean distance or Mahalanobis distance. \cite{mikolov2013efficient}.

Each word in the table, including geological terms and city names, was represented as a 200-dimensional vector. The similarity between the keyword and all other words in the dataset was calculated using the cosine similarity, whose formula is
\[\text{cosine\,similarity}=\frac{u\cdot v}{||u|| ||v||},\]
where $u$ is the vector representing the keyword, and $v$ is the vector representation of any other word in the dataset, providing a similarity score between the keyword and each word in the vocabulary.
After calculating the cosine similarity for all words in the dataset, a filtering step was applied to extract only the city names. This was done using a pre-compiled list of cities derived from Simplemaps \cite{simplemaps}. From the filtered results, the top ten cities with the highest cosine similarity scores were selected for further analysis. These cities were considered to have the strongest semantic relationship with the keyword and were hypothesized to be locations of interest for potential resources.

\subsection{Haversine Equation}
Following the identification of the ten cities most semantically related to the keyword using cosine similarity, the next step in the methodology involved calculating errors by computing their geographical distances to identified lithium mines. This step provides a quantitative measure of how closely the cities predicted by the NLP model align with actual lithium deposits.

We employed the haversine formula to compute the distances between the predicted cities and the nearest lithium mines. The haversine formula calculates the shortest distance between two points on the Earth's surface, given their latitude and longitude, and accounts for the Earth's spherical shape \cite{sinnott1987haversine}. This distance can be calculated as follows.
Let be
\[a = \sin^2\left(\frac{\Delta\phi}{2}\right) + \cos(\phi_1) \cos(\phi_2) \sin^2\left(\frac{\Delta\lambda}{2}\right)\]
\[c = 2 \arctan2\left(\sqrt{a}, \sqrt{1-a}\right)\]
\[d = R \cdot c,\]
where:
\begin{itemize}
\item $\phi_1, \phi_2$ are the latitudes of the two points in radians.
\item $\lambda_1, \lambda_2$ are the longitudes of the two points in radians.
\item $\Delta\phi = \phi_2 - \phi_1$ is the difference in latitude.
\item $\Delta\lambda = \lambda_2 - \lambda_1$ is the difference in longitude.
\item $R$ is the radius of the Earth, whose mean value is R = 6,371 km.
\item $d$ is the great-circle distance between the two points.
\end{itemize}

The result of the haversine equation provides the distance (in kilometers) between each city and the closest lithium mine. These distances were treated as the error in the model’s prediction. Ideally, the closer the predicted city is to a known lithium mine, the better the model's performance in accurately identifying relevant locations based on geological text.

This benchmarking step allowed for the evaluation of the model’s accuracy by comparing the predicted locations (cities) with actual known lithium resources. Any significant discrepancy between predicted cities and actual lithium deposits indicates areas where the model might need further refinement in identifying geologically significant locations. Note that this limitation in our study is because we solely use the cities' names as our predictor, which could be overcome using other predictors such as districts, roads, etc. However, this comes with the potential drawback that such words might not appear in the original texts.
\section{Results}

\subsection{Keyword Selection: ``Lithium''}

The keyword ``lithium'' was chosen due to its increasing importance in resource exploration, particularly for its role in renewable energy technologies \cite{ge2025low,diouf2015potential}. Lithium is critical for battery production, making its discovery a priority in geological research. By focusing on lithium, the study aims to identify cities that have strong semantic relationships to the element based on geological texts.

Using NLP techniques, the word embeddings were analyzed to determine which cities are most closely related to ``lithium'' in the context of the geological corpus. This approach helps to pinpoint regions potentially linked to lithium deposits, providing insights that may support future exploration efforts.

\subsection{Cosine Similarity Analysis}

In this section, we analyze and present the top cities semantically related to the keyword ``lithium,'' identified using cosine similarity, and compare their geographic proximity to known lithium mines using the haversine distance.

To represent the most relevant cities to ``lithium'' based on cosine similarity for each dimensionality reduction method, we created individual tables that list the top ten cities for each method (\Cref{tab:similarity_noreduction,tab:similarity_pca,tab:similarity_ae,tab:similarity_vae,tab:similarity_vae_lstm}). Each table includes the city name and administrative region name (admin-name). This layout highlights how each dimensionality reduction technique produces unique sets of cities with varying levels of similarity to the target keyword, reflecting each method’s interpretative characteristics.

\begin{center}
    \begin{table}
        \centering
        \begin{tabular}{|c|c|c|}
            \hline
            \textbf{Rank} & \textbf{City} & \textbf{Admin-name} \\
            \hline
             1 & manado & Sulawesi Utara \\
             \hline
             2 & laramie & Wyoming \\
             \hline
             3 & cincinnati & Ohio \\
             \hline
             4 & pasco & Washington \\
             \hline
             5 & sherwood & Oregon \\
             \hline
             6 & sherwood & Arkansas \\
             \hline
             7 & wyoming & Ohio \\
             \hline
             8 & wyoming & Michigan \\
             \hline
             9 & alliance & Ohio \\
             \hline
             10 & formosa & Formosa \\
             \hline
        \end{tabular}
        \caption{Baseline cosine similarity results without dimensionality reduction. This table lists the top cities most similar to ``lithium'' in the original, high-dimensional word embedding space. This version captures full contextual relationships but may be less interpretable in terms of clear clustering.}
        \label{tab:similarity_noreduction}
    \end{table}
\end{center}

\begin{center}
    \begin{table}
        \centering
        \begin{tabular}{|c|c|c|}
            \hline
            \textbf{Rank} & \textbf{City} & \textbf{Admin-name} \\
            \hline
             1 & tiffin & Ohio \\
             \hline
             2 & beaufort & South Carolina \\
             \hline
             3 & vashon & Washington \\
             \hline
             4 & metro & Lampung \\
             \hline
             5 & pierre & South Dakota \\
             \hline
             6 & ama & Aichi \\
             \hline
             7 & harrow & Harrow \\
             \hline
             8 & egg & Zürich \\
             \hline
             9 & male & Maale \\
             \hline
             10 & oas & Albay \\
             \hline
        \end{tabular}
        \caption{Cities most related to ``lithium'' by cosine similarity using PCA. As a linear transformation method, PCA emphasizes variance and may capture more general semantic relationships in the dataset. The cosine similarity scores provide a broad view of how cities are semantically linked to ``lithium'' in a linear subspace.}
        \label{tab:similarity_pca}
    \end{table}
\end{center}

\begin{center}
    \begin{table}
        \centering
        \begin{tabular}{|c|c|c|}
            \hline
            \textbf{Rank} & \textbf{City} & \textbf{Admin-name} \\
            \hline
             1 & laval & Quebec \\
             \hline
             2 & laval & Pays de la Loire \\
             \hline
             3 & alvin & Texas \\
             \hline
             4 & trim & Meath \\
             \hline
             5 & moe & Victoria \\
             \hline
             6 & sion & Valais \\
             \hline
             7 & puck & Pomorskie \\
             \hline
             8 & evergreen & Colorado \\
             \hline
             9 & northwood & Hillingdon \\
             \hline
             10 & bellevue & Île-de-France \\
             \hline
        \end{tabular}
        \caption{Cities most related to ``lithium'' by cosine similarity using the Autoencoder. This table shows the results from the autoencoder, a neural network model that captures non-linear relationships within the embedding space. Autoencoder transformations typically enhance clustering for complex datasets, providing a list of cities related to ``lithium'' through learned feature patterns.
}
        \label{tab:similarity_ae}
    \end{table}
\end{center}

\begin{center}
    \begin{table}
        \centering
        \begin{tabular}{|c|c|c|}
            \hline
            \textbf{Rank} & \textbf{City} & \textbf{Admin-name} \\
            \hline
             1 & center & Pennsylvania \\
             \hline
             2 & tota & Couffo \\
             \hline
             3 & westerly & Rhode Island \\
             \hline
             4 & eagle & Idaho \\
             \hline
             5 & phoenix & Arizonia \\
             \hline
             6 & kalispell & Montana \\
             \hline
             7 & pontiac & Michigan \\
             \hline
             8 & pontiac & Illinois \\
             \hline
             9 & albino & Lombardy \\
             \hline
             10 & barrington & Rhode Island \\
             \hline
        \end{tabular}
        \caption{Cities most related to ``lithium'' by cosine similarity using VAE. VAE's probabilistic framework often emphasizes compact, smooth latent spaces. Here, it produces densely clustered cities related to ``lithium,'' with slightly lower separation compared to Autoencoder results, indicative of VAE's tendency to over-cluster.}
        \label{tab:similarity_vae}
    \end{table}
\end{center}

\begin{center}
    \begin{table}
        \centering
        \begin{tabular}{|c|c|c|}
            \hline
            \textbf{Rank} & \textbf{City} & \textbf{Admin-name} \\
            \hline
             1 & apt & Provence-Alpes-Côte d’Azur \\
             \hline
             2 & evans & Colorado \\
             \hline
             3 & evans & Georgia \\
             \hline
             4 & evans & New York \\
             \hline
             5 & moa & Holguín \\
             \hline
             6 & takahashi & Okoyama \\
             \hline
             7 & sand & Vestfold og Telemark \\
             \hline
             8 & fitzgerald & Georgia \\
             \hline
             9 & tut & Adıyaman \\
             \hline
             10 & clifton & Nottingham \\
             \hline
        \end{tabular}
        \caption{Cities most related to ``lithium'' by cosine similarity using VAE-LSTM. Finally, this table captures results from the VAE-LSTM, which combines VAE's probabilistic nature with LSTM's sequence modeling capability. This model tends to distribute cities more evenly within the latent space, providing slight improvements in the differentiation of cities associated with ``lithium.''
}
        \label{tab:similarity_vae_lstm}
    \end{table}
\end{center}

\subsection{Haversine Distance Benchmarking} 
\begin{figure}[ht]
	\centering
    \begin{minipage}[c]{0.5\textwidth}
		\centering
		\includegraphics[width=\textwidth]{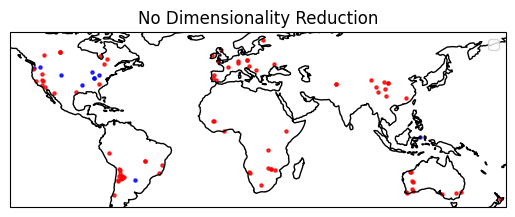}
	\end{minipage}
    \begin{minipage}[c]{0.5\textwidth}
		\centering
		\includegraphics[width=\textwidth]{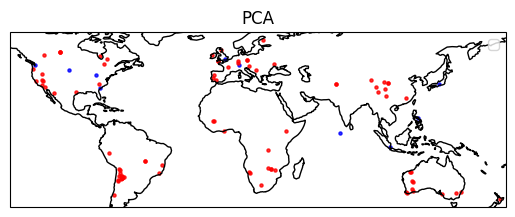}
	\end{minipage}\hfill
	\begin{minipage}[c]{0.5\textwidth}
		\centering
		\includegraphics[width=\textwidth]{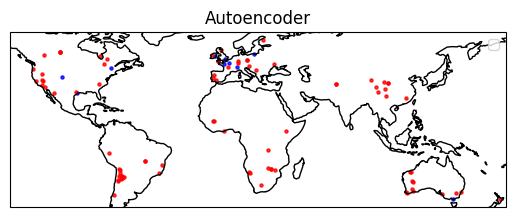}
	\end{minipage}
	\begin{minipage}[c]{0.5\textwidth}
		\centering
		\includegraphics[width=\textwidth]{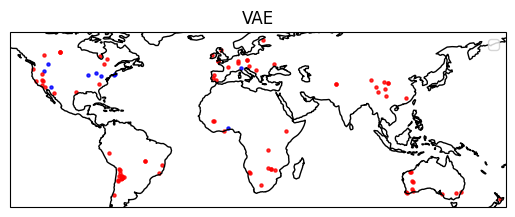}
	\end{minipage}\hfill
    \begin{minipage}[c]{0.5\textwidth}
		\centering
		\includegraphics[width=\textwidth]{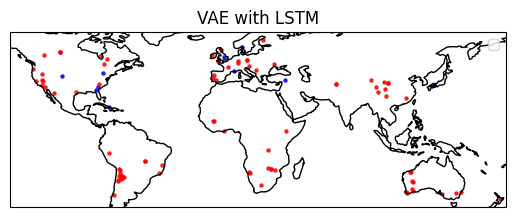}
	\end{minipage}
 	\caption{The figure presents a comparative analysis of five dimensionality reduction techniques—No Reduction, PCA, Autoencoder, VAE, and VAE-LSTM—evaluated through the haversine benchmark. Each method is represented by a world map projection displaying cities selected based on cosine similarity to the keyword ``lithium.'' Cities are plotted as blue dots relative to actual lithium mine locations represented as red dots, which act as a reference for spatial accuracy. The \Cref{tab:rmse_error} summarizes the prediction error for each technique to provide a quantitative measure of distance accuracy. }
	\label{fig:haversine}
\end{figure}

To validate the results of semantic similarity, the geographical proximity of each identified city to known lithium mine locations was calculated using the haversine distance, a metric that calculates the great circle distance between two points on Earth based on latitude and longitude, which are compared in \Cref{tab:rmse_error}. This comparison serves as a reference to assess the real-world relevance of the findings of cosine similarity. \Cref{fig:haversine} offers a visualization of the accuracy of the prediction.

\paragraph{No Dimensionality Reduction}: Serving as a baseline, this projection shows the city-mine distances without applying dimensionality reduction. The cities are more dispersed, with fewer clusters around lithium mine locations. 

\paragraph{PCA:} The PCA projection demonstrates linear clustering due to its reliance on variance maximization. This technique captures only linear relationships, resulting in a broader spread of city points relative to the mines. 

\paragraph{Autoencoder:} The autoencoder effectively captures nonlinear patterns, producing tighter clusters around mining sites. This enhanced clustering shows improved spatial accuracy over PCA and aligns cities more closely with relevant mine locations. 

\paragraph{VAE:} VAE introduces a probabilistic framework, resulting in a compact, densely clustered latent space. This can lead to over-compression, with cities often concentrated centrally, causing overlap. However, the model retains continuity in the latent space, providing reasonably accurate spatial relationships. 

\paragraph{VAE-LSTM:} Incorporating LSTM’s sequence modeling, VAE-LSTM offers a balanced distribution across the latent space. This model reduces the over-compression in VAE alone, producing a diffuse yet coherent clustering that aligns well with lithium mine locations. 

\begin{center}
    \begin{table}
        \centering
        \begin{tabular}{|c|c|c|}
            \hline
            \textbf{Dimensionality Reduction Technique} & \textbf{Prediction Error(km)} \\
            \hline
             No Dimensionality Reduction & 1033.6767\\
             \hline
             PCA & 1662.5537 \\
             \hline
             Autoencoder & 511.8307 \\
             \hline
             Variational Autoencoder(VAE) & 654.9757 \\
             \hline
             VAE with LSTM & 1094.7798 \\
             \hline
             
        \end{tabular}
        \caption{The table above presents the predicted error of the calculated distances for each dimensionality reduction technique, allowing for a direct comparison of spatial accuracy. The prediction error calculated by taking error distance of the first 10 most similar city to do root mean squared error. Lower prediction error values indicate that a technique is more effective at positioning cities close to the lithium mines, as it better preserves the semantic relationships in a way that aligns with the actual locations.
}
        \label{tab:rmse_error}
    \end{table}
\end{center}

\section{Discussion of results}
\label{sec:Discussion}

This study evaluates various dimensionality reduction techniques for mapping semantic relationships in word embeddings to physical proximity, specifically using the keyword ``lithium'' to identify cities with a high likelihood of lithium resources. The primary dimensionality reduction techniques examined were PCA, Autoencoder, VAE, and VAE with LSTM, with results compared based on haversine benchmarking. In this discussion, we analyze the feature extraction performance of each method, as well as the spatial accuracy of the predictions based on the RMSE values derived from the haversine benchmarking. We also consider the broader limitations of the methodology, including challenges with city name ambiguity in cosine similarity scoring.

\subsection{Dimensionality Reduction and Feature Extraction}
Among the four techniques, Autoencoder and VAE-LSTM emerged as the most effective for representing semantic relationships within a 2D latent space. Autoencoder and VAE demonstrated strong non-linear structure preservation, facilitating more distinct clustering of city names relative to other words. VAE-LSTM did not outperform the autoencoder in haversine benchmarking, suggesting that while sequential modeling aids some applications, it may be less beneficial in distinguishing geologically relevant locations.

\subsection{Haversine Benchmarking}
The haversine benchmarking results provide insight into how effectively each dimensionality reduction technique aligns predicted cities with actual lithium mine locations. The autoencoder produced the lowest prediction error, indicating a closer spatial match to known resources. This suggests that non-linear feature extraction methods capture semantic relationships more effectively when mapped to physical proximity.

Higher prediction error in methods like PCA highlight that a simpler linear dimensionality reduction does not capture the complex patterns required for accurate location-based predictions in this domain. Overall, autoencoder and VAE-based techniques outperformed PCA in terms of spatial accuracy, emphasizing the importance of non-linear approaches for geological semantic analysis.

\subsection{Limitations and Future Directions}
The use of cosine similarity to relate cities to the keyword ``lithium'' successfully identified cities with high semantic relevance. However, an observed limitation was the ambiguity in the city names: Different cities sharing the same name were treated as identical in cosine similarity calculations. This limitation caused such cities to receive the same similarity score, potentially skewing the results by grouping semantically unrelated locations. Addressing this issue may involve applying additional location-based data or metadata to distinguish between cities, allowing for more precise semantic mapping.

Another concern is that cities are named after various things, making it difficult to separate them from other words that exist on the plane. This also affects the calculation of cosine similarity since there are times that these city names are mentioned as another meaning that is not correlated to the supposed city. Other characteristic or location indicator could be used in place of cities, but the reason we used it in this research is because it gives enough uniqueness to each of them while containing the problem of confusing names in the manageable level. Giving too specific location keyword means that there is more chance of different places having the same name, which would make the whole system significantly less accurate without proper method to retaliate this problem.

While autoencoder showed strong performance, the model’s dependence on pre-determined embeddings and similarity metrics may not capture all nuances in geological text. Additionally, the observed flaw with city name ambiguity points to a need for more advanced approaches, potentially leveraging additional context or geographic metadata. Future work could explore hybrid models or other dimensionality reduction techniques, such as $t$-SNE \cite{tsne} or UMAP \cite{umap}, which may improve separation without over-compressing relevant features. Addressing these challenges could refine the model’s ability to distinguish geologically significant locations and improve its utility in resource identification applications.
\section{Conclusions}
\label{sec:Conclusions}

This research demonstrates the potential of combining NLP with dimensionality reduction techniques to identify cities semantically related to specific keywords, using geological data as a basis. By leveraging cosine similarity and various dimensionality reduction methods, we assessed the possible location of natural resources by using cities as landmarks. The results indicate that non-linear dimensionality reduction techniques, particularly the autoencoder, enhance the model’s accuracy in mapping semantic relationships to physical proximity, with the autoencoder yielding the lowest RMSE among the methods tested. These findings underscore the capacity of NLP combined with advanced dimensionality reduction to extract meaningful insights from geological text data.

Future research could explore more advanced non-linear models and hybrid approaches, such as combining spatial and semantic embeddings, to further improve prediction accuracy. Moreover, refining the methodology to differentiate cities with shared names and incorporating additional geological factors could enhance the model’s specificity. Broadening the dataset and applying this methodology to diverse resource types would also help assess its generalization, offering promising pathways to support geospatial exploration through NLP.

\section*{Note}
Codes to reproduce our results are available in \href{https://github.com/NanmanasLin/Application-of-natural-language-processing-for-finding-semantic-relation-of-elusive-natural-resource}{https://github.com/NanmanasLin/Application-of-natural-language-processing-for-finding-semantic-relation-of-elusive-natural-resource}

\bibliographystyle{unsrt}
\bibliography{mainBIB}

\end{document}